\documentclass{article}

 \usepackage[dblblindworkshop, final]{neurips_2025}
\workshoptitle{ResponsibleFM}

\usepackage[utf8]{inputenc} 
\usepackage[T1]{fontenc}    
\usepackage{hyperref}       
\usepackage{url}            
\usepackage{booktabs}       
\usepackage{amsfonts}       
\usepackage{nicefrac}       
\usepackage{microtype}      
\usepackage{xcolor}         

\usepackage{subcaption}     

\usepackage{algorithm}
\usepackage{algorithmic}
\usepackage{amsmath,amssymb,amsfonts}
\usepackage{textcomp}
\usepackage{xcolor}
\usepackage{booktabs}
\usepackage{enumitem}
\usepackage[most]{tcolorbox}
\usepackage{wrapfig}        

\title{
    \raisebox{-0.2em}{\includegraphics[height=1.2em]{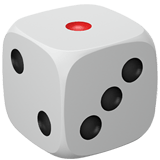}}\;
    DICE: \underline{D}iscrete \underline{I}nterpretable \underline{C}omparative \underline{E}valuation with Probabilistic Scoring for Retrieval-Augmented Generation
}

\author{%
  Shiyan Liu, Jian Ma, Rui Qu \\
  School of Computer Science and Technology \\
  Huazhong University of Science and Technology \\
  Wuhan, China \\
  \texttt{\{shyl, fy47, rqu\}@hust.edu.cn} \\
}

\begin{document}
\maketitle

\begin{abstract}
As Retrieval-Augmented Generation (RAG) systems evolve toward more sophisticated architectures, ensuring their trustworthiness through explainable and robust evaluation becomes critical. Existing scalar metrics suffer from limited interpretability, inadequate uncertainty quantification, and computational inefficiency in multi-system comparisons, hindering responsible deployment of RAG technologies. We introduce DICE (Discrete Interpretable Comparative Evaluation), a two-stage, evidence-coupled framework that advances explainability and robustness in RAG evaluation. DICE combines deep analytical reasoning with probabilistic $\{A, B, Tie\}$ scoring to produce transparent, confidence-aware judgments that support accountable system improvement through interpretable reasoning traces, enabling systematic error diagnosis and actionable insights. To address efficiency challenges at scale, DICE employs a Swiss-system tournament that reduces computational complexity from $O(N^2)$ to $O(N \log N)$, achieving a 42.9\% reduction in our eight-system evaluation while preserving ranking fidelity. Validation on a curated Chinese financial QA dataset demonstrates that DICE achieves 85.7\% agreement with human experts, substantially outperforming existing LLM-based metrics such as RAGAS. Our results establish DICE as a responsible, explainable, and efficient paradigm for trustworthy RAG system assessment.
\end{abstract}

\section{Introduction}

Retrieval-Augmented Generation (RAG) systems represent an emerging paradigm at the intersection of information retrieval and advanced AI technologies~\citep{lewis2020retrieval}. By enabling large language models (LLMs) to retrieve and reason over external corpora, RAG addresses fundamental limitations of purely parametric models, including knowledge cutoffs, hallucinations, and inability to incorporate rapidly evolving information. As RAG systems are increasingly deployed in high-stakes domains such as finance, law, healthcare, and scientific research, ensuring their trustworthiness through explainable, robust, and efficient evaluation becomes paramount~\citep{gao2023retrieval}.

Despite the growing adoption of RAG systems, evaluating these systems remains challenging, particularly regarding explainability and robustness. Traditional text generation metrics such as BLEU~\citep{papineni2002bleu} and ROUGE~\citep{lin2004rouge} focus on surface-level overlap and fail to capture semantic adequacy. Embedding-based metrics like BERTScore~\citep{zhangbertscore} and MoverScore~\citep{zhao2019moverscore}, as well as learned metrics such as BLEURT~\citep{sellam2020bleurt} and COMET~\citep{rei2020comet}, provide stronger semantic correlation but remain poorly suited to evidence-coupled tasks. Distribution-based measures such as MAUVE~\citep{pillutla2021mauve} are primarily designed for open-ended generation and may not be optimal for factuality-sensitive evaluation.

Recent advances in reference-free evaluation, such as BARTScore~\citep{yuan2021bartscore}, GPTScore~\citep{fu2024gptscore}, and SelfCheckGPT~\citep{manakul2023selfcheckgpt}, attempt to estimate quality or factuality without human-written references. For RAG systems in particular, RAGAS (Retrieval Augmented Generation Assessment)~\citep{es2024ragas} introduced multi-dimensional scalar metrics that assess faithfulness, answer relevance, and context utilization. However, scalar scores often obscure meaningful differences between systems and provide limited explainability for responsible and trustworthy deployment. LLM-as-a-judge approaches~\citep{zheng2023judging, chiang2024chatbot} show promising alignment with human preferences but suffer from robustness concerns including systematic biases~\citep{wang2024large, saito2023verbosity}, prompt sensitivity~\citep{lu2024prompts}, and lack of uncertainty quantification. Crucially, existing scalar metrics and LLM-based judgments rarely explain why one system is rated higher than another, hindering transparent and accountable system improvement.

To address these challenges, we introduce DICE\footnote{Code available at \url{https://github.com/shiyan-liu/DICE}} (Discrete Interpretable Comparative Evaluation), a framework that advances explainability, robustness, and efficiency in RAG system evaluation. DICE employs a two-stage, evidence-coupled evaluation process: the first stage grounds each judgment in retrieved context through deep analytical reasoning, enhancing explainability through transparent reasoning traces; the second stage produces probabilistic confidence-aware scores, translating qualitative $\{A, B, Tie\}$ judgments into quantitative measures that capture uncertainty and improve robustness. By combining discrete comparison with probability-based scoring, DICE enables principled and interpretable evaluation while supporting robust statistical comparisons. Furthermore, to address efficiency challenges in the era of neural information retrieval, DICE incorporates a Swiss-system tournament~\citep{csato2013ranking} to scale multi-system ranking, significantly reducing computational complexity without sacrificing fidelity.

We validate DICE on a challenging Chinese financial QA (Question-Answer) dataset consisting of 70 carefully curated QA pairs derived from authoritative sources. Eight RAG systems covering diverse embedding, chunking, and generation strategies are benchmarked. Experiments show that DICE achieves higher agreement with human experts than scalar metrics, provides robust confidence intervals, and identifies nuanced failure modes through interpretable error diagnostics, offering actionable insights for responsible system improvement. 

Our contributions align with workshop themes of defining and measuring trustworthiness, standardized and reproducible evaluation protocols, and datasets \& benchmarks for responsible deployment:

\begin{itemize}
\item  We propose DICE, a two-stage, evidence-coupled evaluation framework that enhances transparency and accountability through interpretable reasoning traces and probabilistic, confidence-aware scoring, enabling measurement of truthfulness and robust, evidence-grounded $\{A, B, Tie\}$ judgments to support accountable system improvement.
\item We operationalize a standardized, reproducible evaluation protocol for RAG systems, specifying judge prompts, decision vocabularies, scoring rules, and tournament procedures, enabling consistent comparisons across systems and studies.
\item We construct a challenging Chinese financial QA dataset and benchmark eight diverse RAG systems, achieving 85.7\% accuracy and a Cohen’s $\kappa$ of 0.742 ~\citep{cohen1960coefficient}, substantially outperforming existing LLM-based metrics such as RAGAS. We emphasize ethical curation and transparent documentation practices to facilitate responsible benchmarking and trustworthy deployment.
\end{itemize}

\section{Related Work}

\subsection{Automated Evaluation of Text Generation Systems}

Automated evaluation of text generation has evolved from surface-level lexical overlap metrics to semantic and model-based approaches. Early metrics such as BLEU~\citep{papineni2002bleu} and ROUGE~\citep{lin2004rouge} remain widely used but are limited in capturing semantic adequacy and lack explainability regarding their scoring decisions. Embedding-based measures such as BERTScore~\citep{zhangbertscore} and MoverScore~\citep{zhao2019moverscore} improved correlation with human judgments by leveraging contextualized embeddings. Learned metrics, including BLEURT~\citep{sellam2020bleurt} and COMET~\citep{rei2020comet}, train neural evaluators on human-annotated or synthetic data, further advancing alignment with human preferences. However, these neural approaches introduce concerns about robustness against distribution shifts and interpretability of their internal decision processes.

More recently, reference-free methods have become popular, with BARTScore~\citep{yuan2021bartscore} reframing evaluation as conditional likelihood estimation and GPTScore~\citep{fu2024gptscore} demonstrating that prompting large models directly can yield reliable multi-aspect scores. Recent work on factuality and hallucination detection has introduced sophisticated approaches including SelfCheckGPT~\citep{manakul2023selfcheckgpt}, which detects inconsistencies by sampling multiple outputs from the same model, and token-probability-based methods that exploit model confidence for quality estimation. These approaches complement traditional metrics by focusing specifically on factual accuracy and reliability, which are crucial for knowledge-intensive applications. However, they remain vulnerable to systematic biases and lack transparency in their evaluation rationale, limiting their applicability for responsible and trustworthy deployment. In addition, standardized and reproducible evaluation protocols are often under-specified (e.g., prompts, decision vocabularies, and aggregation rules), hindering consistent comparisons across studies; DICE explicitly specifies these elements to promote rigor and comparability.

For RAG systems representing emerging paradigms at the intersection of retrieval and generation, dedicated frameworks such as RAGAS~\citep{es2024ragas} propose evaluation across faithfulness, answer relevance, and context utilization, emphasizing evidence-coupled assessment. However, these approaches often lack explainability and uncertainty quantification essential for accountable system improvement. DICE addresses these critical limitations by combining discrete pairwise comparisons with probabilistic, confidence-aware scoring. This design ensures explainable, robustness-aware, and uncertainty-quantified judgments, particularly important for responsible deployment of RAG systems that require retrieval-grounded reasoning and evidence-based validation.

\subsection{Pairwise Judgment through LLMs}

Pairwise evaluation has emerged as a complementary paradigm, leveraging the intuitive nature of comparative judgments over absolute scoring. This approach underpins reinforcement learning from human feedback (RLHF)~\citep{ouyang2022training} where preference data provides more consistent supervision than scalar ratings. Recent work has successfully adapted pairwise evaluation to diverse NLP tasks, demonstrating superior reliability to Likert-scale assessments~\citep{stiennon2020learning}.

~\citet{zheng2023judging} established foundational principles for LLM-based pairwise evaluation, showing that \textsc{gpt-4}~\citep{achiam2023gpt} can provide robust comparative judgments across text generation tasks. Chatbot Arena~\citep{chiang2024chatbot} scaled this paradigm to large-scale conversational AI evaluation through crowdsourced preferences. These systems often employ Elo rating systems to maintain dynamic rankings and enable efficient tournament-style evaluation structures. Despite their effectiveness, pairwise approaches face critical challenges regarding explainability, robustness, and efficiency when applied to RAG systems: most lack transparent reasoning traces explaining why one system outperforms another, remain vulnerable to position bias and inconsistent judgments, and become computationally infeasible through exhaustive tournaments when evaluating many systems.

For trustworthy and responsible RAG deployment, DICE adapts pairwise evaluation to retrieval-grounded settings by coupling judgments with retrieved evidence, quantifying confidence over \{A, B, Tie\} decisions, and organizing multi-system comparison efficiently. At a high level, DICE provides an evidence-aware, confidence-calibrated, and scalable evaluation procedure for RAG, without relying on task-specific heuristics, enabling responsible and comparable assessments in the neural information retrieval era.

\section{The DICE Framework}

\begin{figure*}[!t]
\centering
\includegraphics[width=1\linewidth]{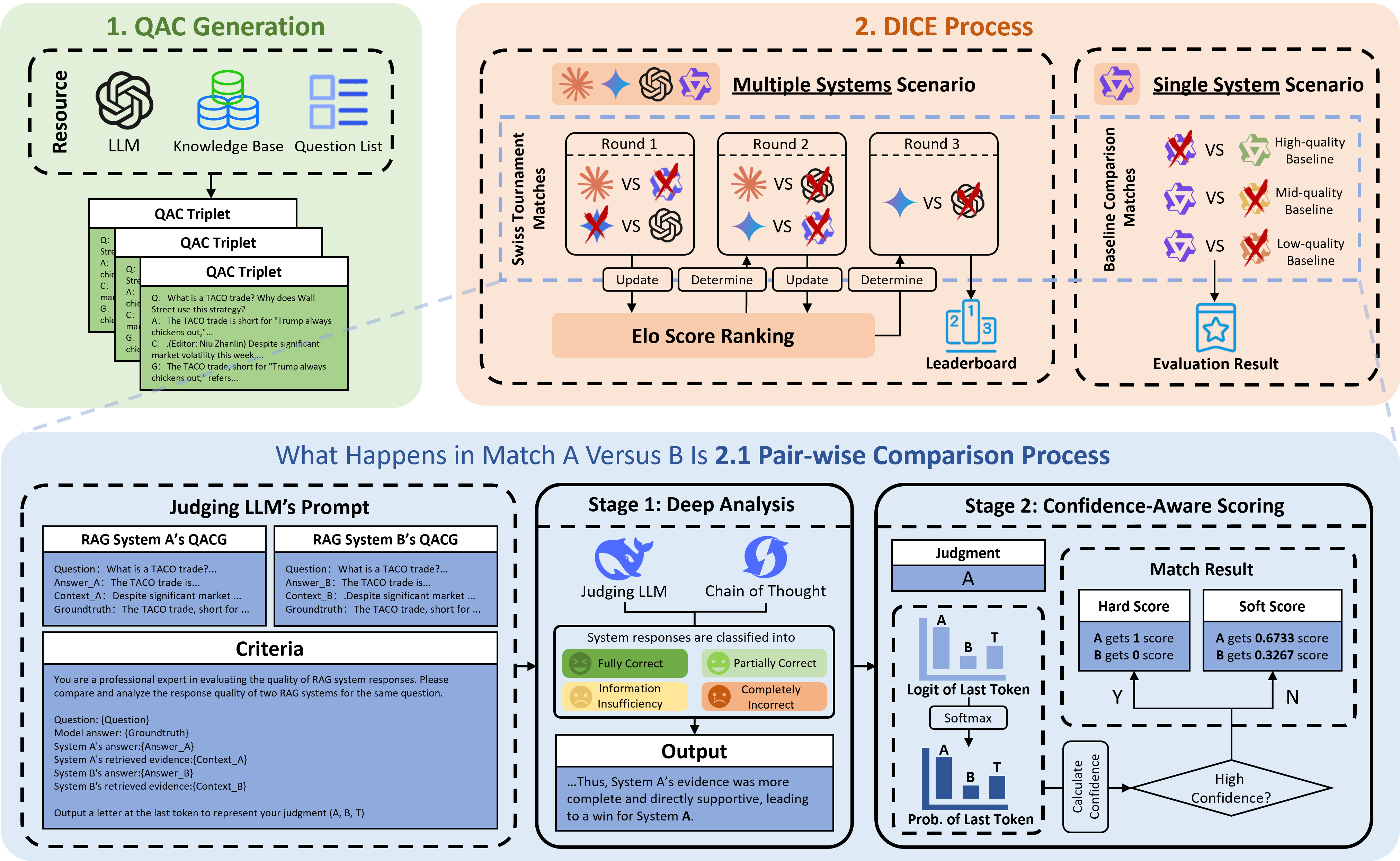}
\caption{Overview of the DICE framework: the complete workflow from data preparation to multi-system ranking via Swiss-system tournament and single-system baseline comparison, along with a concrete example of the two-stage evaluation process.}
\label{fig:framework_overview}
\end{figure*}

DICE provides a comprehensive evaluation framework addressing interpretability, uncertainty quantification, and computational efficiency challenges. Figure \ref{fig:framework_overview} illustrates the complete workflow, including overall architecture and concrete evaluation examples. The framework operates through two stages separating analytical reasoning from probabilistic scoring.

\subsection{Stage I: Evidence-Coupled Deep Analysis}

DICE evaluates RAG systems using a two-stage, evidence-coupled protocol. The first stage produces detailed reasoning and analysis of each system output, while the second stage generates a final probabilistic decision that accounts for uncertainty. By separating thorough analysis from decision-making, DICE ensures both transparent evaluation and reliable, interpretable judgments.

In the first stage, the evaluator examines each evidence-answer pair holistically, considering factual accuracy relative to ground-truth information, completeness in addressing all aspects of the query, and the quality of evidence integration with retrieved documents. This deep analysis generates structured reasoning traces that support systematic evaluation, consistency verification, and downstream error analysis. These reasoning traces provide the foundation for interpretable evaluation, allowing developers to understand specific failure modes and areas for improvement.

Following the analysis, a single-token judgment is produced from the vocabulary $\{A, B, Tie\}$, guided by task-specific ranking criteria. System responses are classified into four hierarchical categories:

\begin{description}[leftmargin=0em, labelwidth=10em]
\item[Fully correct answers] Responses that are accurate and integrate all relevant evidence; ranked highest.
\item[Partially correct answers] Responses that address some but not all aspects of the question; ranked below fully correct answers.
\item[Information insufficiency] Responses that explicitly acknowledge a lack of knowledge; ranked below partially correct answers.
\item[Completely incorrect answers] Responses that contradict canonical references or provide misleading information; ranked lowest.
\end{description}

In pairwise comparisons, this hierarchy ensures systematic ranking: fully correct answers win over all other types; partially correct answers win over insufficient or incorrect answers but lose to fully correct ones; insufficient answers win only over completely incorrect answers; and completely incorrect answers lose to all other types. The pairwise judgments produced through this systematic analysis help developers understand why one system outperforms another, providing actionable insights for model improvement rather than opaque numerical scores.

\subsection{Stage II: Probabilistic Confidence-Aware Scoring}

The second stage of DICE builds on the deep analysis produced in the first stage and implements a probabilistic confidence-aware scoring mechanism to translate the qualitative evaluation into quantitative, pairwise scores. After generating a single-token decision from the vocabulary $\{A, B, Tie\}$, logits for each token are extracted and converted into a probability distribution using the softmax function. Let $l_A$, $l_B$, and $l_{Tie}$ denote the logits for tokens A, B, and Tie respectively. The probability distribution $P$ is computed as:
\begin{equation}
P = \text{softmax}([l_A, l_B, l_{Tie}]) = \frac{e^{l_i}}{\sum_{j \in \{A, B, Tie\}} e^{l_j}}
\end{equation}
To quantify the confidence of the judgment, we compute the margin between the highest and second-highest probabilities. Let $P_{\max}$ represent the maximum probability among all judgment tokens and $P_{\text{second}}$ denote the second-highest probability. The probability margin $\Delta P$ is then calculated as:
\begin{equation}
\Delta P = P_{\max} - P_{\text{second}}
\end{equation}
This margin determines whether the judgment is treated as high-confidence or low-confidence. The confidence threshold of 0.1 was empirically determined through grid search on a validation set to optimize agreement with human expert judgments.

For high-confidence cases where $\Delta P \geq 0.1$, we apply a hard scoring scheme $S_{\text{hard}}$ that assigns a decisive outcome:
\begin{equation}
S_{\text{hard}} =
\begin{cases}
(1,0), & \text{if } \arg\max(P) = A \\
(0,1), & \text{if } \arg\max(P) = B \\
(0.5,0.5), & \text{if } \arg\max(P) = Tie
\end{cases}
\end{equation}
For low-confidence cases where $\Delta P < 0.1$, we employ a soft scoring scheme $S_{\text{soft}}$ that redistributes the tie probability proportionally between A and B. Let $p_A$, $p_B$, and $p_{\text{Tie}}$ denote the individual probabilities for tokens A, B, and Tie respectively. The soft scoring is computed as:
\begin{equation}
\begin{aligned}
S_{\text{soft}} &= \left(
    p_A + p_{\text{Tie}} \cdot \frac{p_A}{p_A + p_B},\;
    p_B + p_{\text{Tie}} \cdot \frac{p_B}{p_A + p_B}
  \right)
\end{aligned}
\end{equation}
By explicitly separating the probabilistic scoring from the deep analysis of the first stage, this second stage captures graded evaluation uncertainty while maintaining interpretability. Together with the first stage, this probabilistic scoring mechanism enables DICE to produce transparent, reliable, and confidence-aware pairwise assessments of knowledge-intensive RAG systems.
\begin{algorithm}[!t]
\caption{Swiss-System Tournament for RAG Evaluation}
\label{alg:swiss_tournament}
\begin{algorithmic}[1]
\REQUIRE Systems $S = \{s_1, s_2, ..., s_N\}$, Questions $Q$, Rounds $R$
\STATE Initialize Elo ratings: $\text{rating}[s_i] = 1500$ for all $s_i \in S$
\FOR{round $r = 1$ to $R$}
    \STATE Pair systems by current Elo while avoiding repeat matches
    \FOR{each pair $(s_i, s_j)$}
        \STATE Execute pairwise comparisons with confidence-aware scoring
        \STATE Compute cumulative scores $S_{total}^i, S_{total}^j$
        \STATE Update Elo ratings using weighted formulation
    \ENDFOR
\ENDFOR
\RETURN Final Elo rankings
\end{algorithmic}
\end{algorithm}

\subsection{System Evaluation Framework}

DICE provides a unified evaluation framework supporting both multi-system tournaments and single-system baseline comparisons. The framework ensures interpretable ranking, efficient computation, and relative performance assessment, accommodating both large-scale system comparisons and focused individual system analysis.

\paragraph{Multi-System Tournament} 
For multi-system evaluation, DICE employs a Swiss-system tournament~\citep{csato2013ranking} to reduce computational complexity from $O(N^2)$ for exhaustive pairwise evaluation to $O(N \log N)$. This approach significantly improves efficiency compared to round-robin evaluation: for eight systems, four rounds of four matches each yield 16 comparisons versus 28 in a full round-robin tournament, representing a 42.9\% reduction in computational cost.

Systems are initially assigned equal Elo ratings and paired dynamically each round based on current performance estimates while avoiding repeat matches. The pairing strategy prioritizes systems with similar Elo scores to maximize information gain while maintaining competitive balance, thereby preserving ranking fidelity. Elo ratings are updated after each match using cumulative soft win scores. For a system A with current rating $R_A$ competing against system B with rating $R_B$, the updated rating $R'_A$ is calculated as:
\begin{equation}
R'_A = R_A + K \cdot \left(\frac{S_{total}^A}{N_{questions}} - E_A\right)
\end{equation}
where $K$ is the K-factor determining the magnitude of rating updates, $S_{total}^A$ represents the cumulative score for system A defined as $S_{total}^A = \sum_{i=1}^{N_{questions}} S_i^A$ across all questions in a match, with $S_i^A$ denoting the score assigned to system A for question $i$, $N_{questions}$ is the total number of questions evaluated, and $E_A$ is the expected score following the standard Elo formulation:
\begin{equation}
E_A = \frac{1}{1 + 10^{(R_B - R_A)/400}}
\end{equation}
Weighted Elo updates incorporate rating differences and upset bonuses, amplifying K-factors when lower-rated systems outperform higher-rated opponents to accelerate convergence and improve ranking accuracy. The complete Swiss-system tournament procedure is outlined in Algorithm~\ref{alg:swiss_tournament}.

\paragraph{Single-System Baseline Evaluation}
For targeted assessment of a single system, DICE provides a baseline evaluation mode to contextualize its performance. This mode utilizes a set of representative baselines to establish High, Medium, and Low-quality performance tiers. These baselines are actual RAG systems selected from the results of previously-run tournaments, chosen for their consistent demonstration of top-tier, mid-tier, or low-tier performance. To evaluate a new target system, it is compared against the pre-generated answer sets from each of these three representative baseline systems using the identical pairwise comparison protocol. This process yields quantitative statistics, such as win counts against each quality tier and a final Elo score. By benchmarking against real, established systems, this approach transforms an absolute measurement into a meaningful relative positioning, allowing practitioners to interpret the system's capabilities within a well-defined quality spectrum.

\section{Experiments}

\subsection{Dataset}

We constructed a challenging Chinese financial QA dataset to validate DICE. The dataset comprises 70 QA pairs curated from real-world financial news articles and commentaries from major Chinese media outlets. QA pairs were generated via a large language model to ensure they are grounded in the source documents, which cover diverse topics like market analysis, corporate performance, and regulatory events. Curation followed ethical and transparent documentation practices (e.g., recording data provenance, source attributions, and selection criteria) to support responsible benchmarking.

The dataset is designed to test a range of cognitive skills essential for financial analysis, including factual recall, multi-hop reasoning, numerical computation, and temporal reasoning. The difficulty spans from basic financial literacy to expert-level analysis, ensuring the evaluation is comprehensive and relevant to real-world applications.

\subsection{Experimental Setup}

All DICE evaluations were conducted using DeepSeek-R1~\citep{guo2025deepseek} as the underlying judge model, chosen for its strong reasoning capabilities and multilingual performance on Chinese financial content.

We evaluated DICE under two complementary scenarios: (i) human expert validation on a curated test set, and (ii) a Swiss-system tournament with multiple RAG systems. This setup enables assessment of system performance under both absolute and relative evaluation conditions, providing comprehensive validation of the framework's effectiveness.

\paragraph{Reproducibility Protocol}
To promote standardized and reproducible evaluation, we fix judge model versions and decoding parameters, publish the decision vocabulary ($\{A, B, Tie\}$) and scoring rules (hard/soft schemes), and share the exact judgment prompts and pairing/tournament procedures. We also control random seeds for sampling-based components and log probabilities for confidence-aware scoring to ensure determinism where possible.

\paragraph{Human Expert Validation}
Two RAG systems--S1 and S8 from Table~\ref{tab:systems}--were selected as System A and System B respectively to provide meaningful performance variation for sensitivity analysis. Three financial domain professionals, each with advanced degrees and at least five years of experience, independently evaluated 70 QA pairs using standardized rubrics emphasizing factual accuracy, completeness, and evidence integration. Disagreements were resolved via majority vote, with tie cases addressed through discussion and consensus.

\begin{wraptable}{r}{0.55\linewidth}
    \vspace{-\baselineskip}
    \centering
    \caption{Eight RAG system configurations evaluated in Swiss-system tournament, arranged in order of decreasing architectural complexity across embedding models, chunking strategies, and generation capabilities.}
    \label{tab:systems}
    \begin{tabular}{llcl}
    \toprule[0.13em]
    \textbf{System} & \textbf{Embedding} & \textbf{Chunk} & \textbf{Gen Model} \\
    \midrule
    S1 & bge-small-zh & 256 & Qwen2.5-7B \\
    S2 & bge-small-zh & 512 & Qwen2.5-7B \\
    S3 & bge-large-zh & 256 & Qwen2.5-7B \\
    S4 & bge-large-zh & 512 & Qwen2.5-7B \\
    S5 & bge-small-zh & 256 & Qwen2.5-0.5B \\
    S6 & bge-small-zh & 512 & Qwen2.5-0.5B \\
    S7 & bge-large-zh & 256 & Qwen2.5-0.5B \\
    S8 & bge-large-zh & 512 & Qwen2.5-0.5B \\
    \bottomrule[0.13em]
    \end{tabular}
\end{wraptable}

Evaluation against the labeled test set employed the RAGAS framework~\citep{es2024ragas}, which assesses RAG systems across three core dimensions established in the original framework: Faithfulness (measuring whether generated answers are grounded in the provided context without hallucinations), Answer Relevancy (evaluating how directly and completely the generated answer addresses the user's question), and Context Relevance (assessing whether the retrieved context contains minimal irrelevant information). Following the original RAGAS methodology, the total score for a system $S$ on a question $q$ is computed as the arithmetic mean of the three dimension scores $d_i(S,q)$ for $i \in \{1,2,3\}$:
\begin{equation}
\label{eq:score_s}
\text{Score}_S(q) = \frac{1}{3} \sum_{i=1}^{3} d_i(S,q)
\end{equation}
Here, $d_i(S,q)$ represents the normalized evaluation score for dimension $i$ corresponding to Faithfulness, Answer Relevancy, and Context Relevance respectively. This unweighted averaging approach aligns with the original RAGAS paper, which does not prescribe specific weighting schemes for combining these fundamental evaluation dimensions. Pairwise comparison labels $\{A, B, Tie\}$ were derived by comparing RAGAS scores with a threshold $\Delta = 0.15$, consistent with the procedure applied across human expert judgments.

\paragraph{Swiss-System Tournament Evaluation}
Eight RAG systems were evaluated in a $2 \times 2 \times 2$ factorial design across embedding models (bge-small-zh vs. bge-large-zh~\citep{xiao2024c}), chunk sizes (256 vs. 512 tokens), and generation models (Qwen2.5-7B vs. Qwen2.5-0.5B~\citep{qwen2025qwen25technicalreport}). Table~\ref{tab:systems} summarizes the system configurations arranged by decremental complexity.

The tournament was conducted over four rounds with four matches per round, totaling 16 comparisons versus 28 for exhaustive pairwise evaluation, demonstrating the efficiency gains of the Swiss-system approach. Systems were initialized with equal Elo ratings and paired dynamically based on current performance while avoiding repeat matchups. Elo ratings were updated after each round. All systems accessed identical document corpora, applied standardized input formatting, and performed retrieval independently. Generation was restricted to retrieved context, preventing external information influence. For the RAGAS baseline in the tournament, pairwise $\{A, B, Tie\}$ labels were derived from the weighted scoring formula in Equation \eqref{eq:score_s} with the same threshold methodology.

\subsection{Results}

\paragraph{Human Expert Validation}  

DICE was evaluated on 70 Chinese financial QA pairs against three human experts. Table~\ref{tab:human_validation} summarizes agreement metrics for DICE and the RAGAS baseline, where RAGAS scores were converted to $\{A, B, Tie\}$ labels using the threshold-based approach.

\begin{wraptable}{r}{0.45\linewidth}
\centering
\caption{Agreement metrics between automated evaluation systems and human expert judgments on Chinese financial QA dataset.}
\label{tab:human_validation}
\begin{tabular}{lcc}
\toprule[0.13em]
\rule{0pt}{10pt}
\textbf{System} & \textbf{Accuracy (\%)} & \textbf{Cohen's $\kappa$} \\
\midrule
DICE & 85.7 & 0.742 \\
RAGAS & 45.7 & 0.096 \\
\bottomrule[0.13em]
\rule{0pt}{5pt}
\end{tabular}
\end{wraptable}

DICE achieves substantially higher agreement with human experts than RAGAS, with 85.7\% accuracy representing a significant improvement over the 45.7\% baseline. The Cohen's kappa~\citep{cohen1960coefficient} value of 0.742 indicates substantial agreement beyond chance, demonstrating that DICE's two-stage evidence-coupled evaluation process successfully captures nuanced performance differences that align with professional financial expertise. This level of robustness is particularly critical for trustworthy RAG deployment in high-stakes financial applications, where subtle differences in factual accuracy and evidence interpretation can have significant practical consequences.

\begin{wrapfigure}{r}{0.45\linewidth}
\centering
\includegraphics[width=\linewidth]{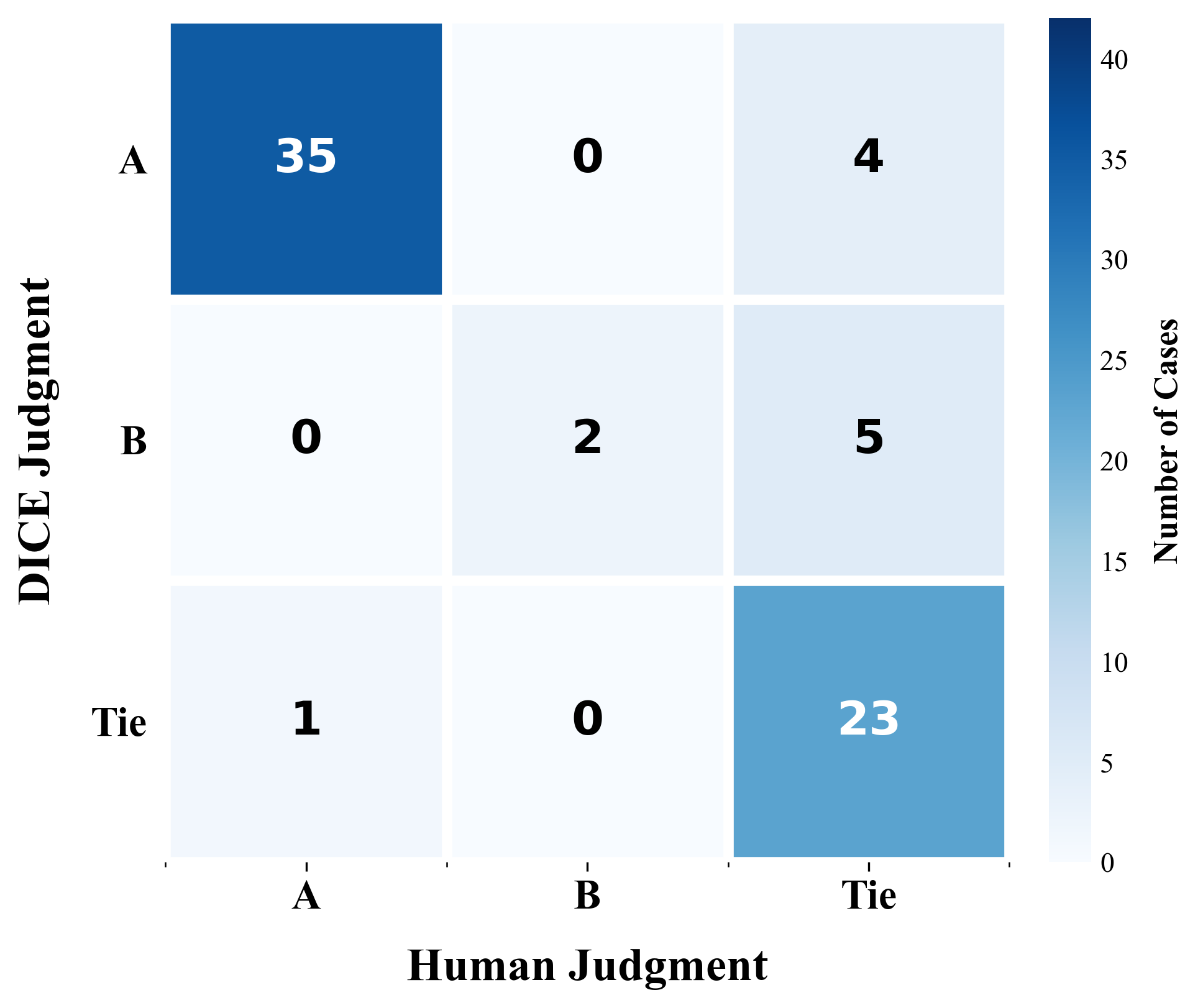}
\caption{Confusion matrix detailing alignment patterns between DICE judgments and human expert consensus.}
\label{fig:results_confusion}
\vspace{-2em}
\end{wrapfigure}

Figure~\ref{fig:results_confusion} presents the detailed confusion matrix between DICE and human expert judgments, illustrating the distribution of agreement and disagreement patterns across all three judgment categories $\{A, B, Tie\}$.

Analysis of the confusion matrix reveals that the remaining disagreements between DICE and experts were concentrated in borderline cases where DICE applied more stringent evaluation criteria than human evaluators. This pattern suggests that DICE may be more conservative in tie classifications, potentially reflecting the deterministic nature of automated evaluation versus the contextual, experience-driven judgments of financial domain experts.

\begin{figure*}
\centering
\includegraphics[width=\linewidth]{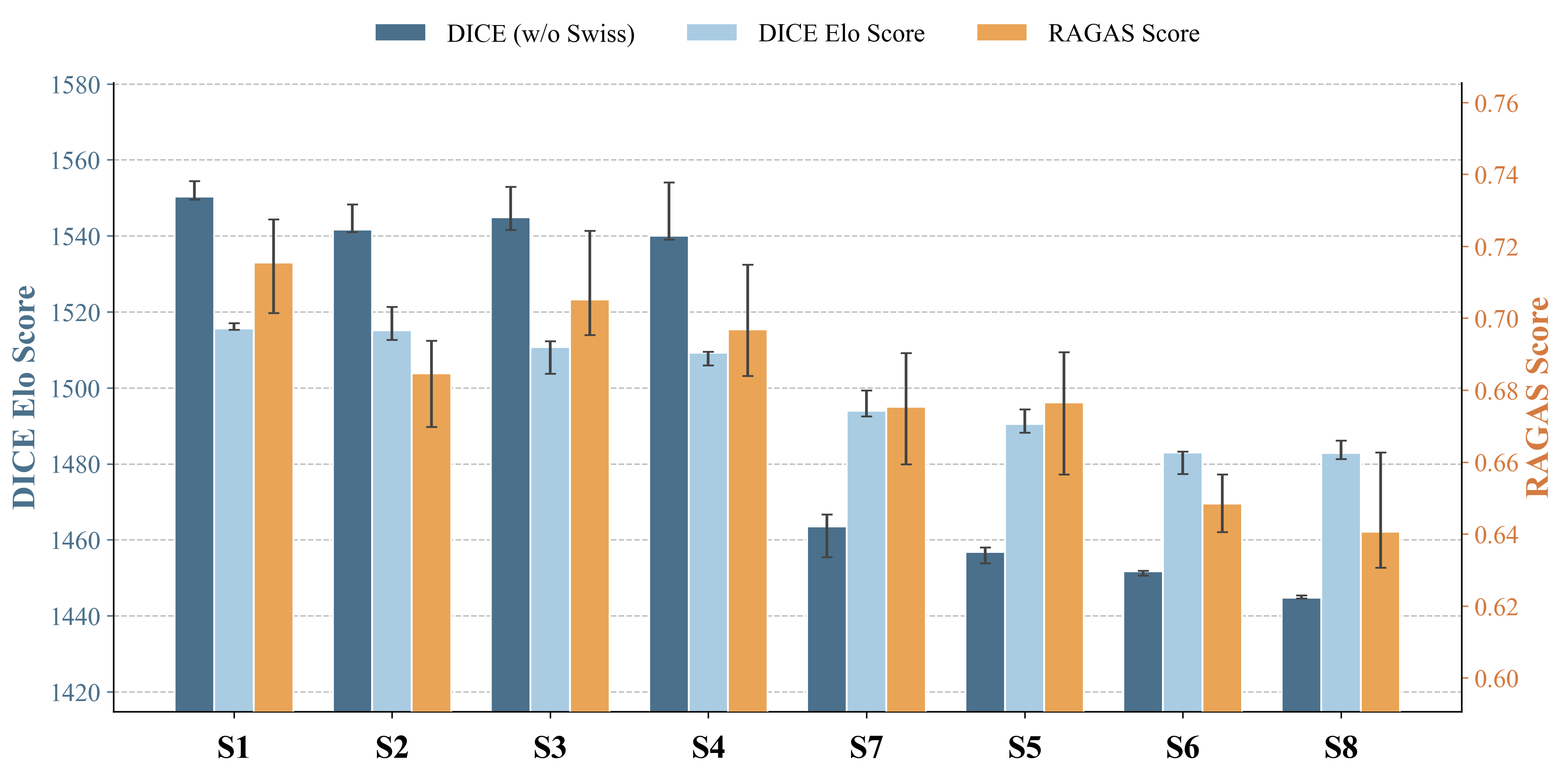}
\caption{Final Elo rankings comparing DICE and RAGAS across eight RAG configurations.}
\label{fig:results_elo}
\end{figure*}

\paragraph{Swiss-System Tournament Results}  
Eight RAG systems were evaluated in a four-round Swiss-system tournament with 16 total matches. Final Elo rankings for DICE and the RAGAS baseline are presented in Figure~\ref{fig:results_elo}. The figure includes results from both the Swiss-system tournament (16 comparisons) and the complete exhaustive pairwise evaluation (w/o Swiss, 28 comparisons), where the latter represents the full version with all pairwise comparisons conducted.

The tournament reveals distinct performance patterns and notable ranking differences between DICE and RAGAS evaluation frameworks. Under DICE evaluation, systems demonstrate a clear performance hierarchy with generation model capacity emerging as the dominant factor: systems with Qwen2.5-7B (S1--S4) consistently outperform those with Qwen2.5-0.5B (S5--S8). Within each generation model tier, both embedding model and chunk size configurations provide incremental performance variations, with S1 (bge-small-zh, 256 tokens, Qwen2.5-7B) ranking highest.

RAGAS produces a different ranking pattern that places greater emphasis on specific configuration combinations rather than consistent architectural progression. Notably, RAGAS shows different sensitivities to embedding model choices within the same generation model tier, suggesting that scalar metrics may capture different aspects of system performance compared to DICE's evidence-coupled comparative approach.

Validation against the exhaustive baseline (w/o Swiss) confirms identical rankings across all systems. Crucially, the Swiss-system demonstrates minimal variance with shorter error bars than the exhaustive approach, achieving superior precision and stability while reducing computational cost by 42.9\%.

These ranking differences provide important insights for practical system deployment. DICE's clear separation between 7B and 0.5B model tiers suggests that practitioners should prioritize larger generation models for financial QA applications, with embedding and chunking optimizations providing secondary benefits. The divergent rankings between evaluation frameworks demonstrate that DICE captures distinct evaluation perspectives, potentially offering more reliable guidance for evidence-intensive applications where factual accuracy and logical coherence are paramount.

\section{Conclusion}

We propose DICE, a two-stage framework that advances the trustworthy evaluation of RAG systems by uniting evidence-coupled analysis, confidence-aware \(\{A, B, Tie\}\) scoring, and a standardized, reproducible protocol. On a Chinese financial QA benchmark, it achieves 85.7\% accuracy and a Cohen’s \(\kappa\) of 0.742, while the Swiss-system tournament ensures \(O(N \log N)\) scalability. The framework provides interpretable reasoning traces, probability-grounded robustness signals, and efficient, high-fidelity ranking—all essential for reliable deployment in high-stakes settings. DICE therefore offers a responsible, explainable, and scalable paradigm aligned with emerging standards for trustworthiness measurement and ethically curated benchmarks.

\section{Limitations}

While DICE demonstrates strong empirical performance, several limitations remain across trustworthiness, robustness, and explainability. Our 70-QA, single-domain (Chinese finance) evaluation constrains out-of-distribution generalization; thresholds and scoring may require domain-specific calibration; the judge model can introduce bias or be vulnerable to adversarial inputs; and reasoning traces may be incomplete on edge cases. Future work will extend evaluation to broader domains and system cohorts, incorporate bias audits and privacy-preserving retrieval with red-teaming, and develop adaptive calibration and ensemble strategies to further enhance robustness while preserving explainability and efficiency.

\section*{Acknowledgements}
The first author acknowledges JD.com for the opportunity to work with real-world RAG systems during a prior internship, where practical challenges encountered in evaluation inspired aspects of this research direction. The work presented in this paper was conducted independently at Huazhong University of Science and Technology after the internship concluded. We are grateful to the anonymous reviewers for their constructive feedback that helped improve this manuscript.

\bibliographystyle{plainnat}
\bibliography{neurips_2025}

\appendix

\onecolumn        
\clearpage        

\section{DICE Evaluation Process Example}

\tcbset{
    colback=gray!5,
    colframe=black!60,
    boxrule=1pt,
    arc=1mm,
    breakable,
    left=4mm, right=4mm, top=3mm, bottom=3mm
}

\begin{tcolorbox}
{\large\textbf{Question:}}\par\medskip
What is a TACO trade? Why does Wall Street use this strategy?\par\bigskip

{\large\textbf{System A Response:}}\par\medskip
{\textbf{Answer:}}\par
TACO trading involves the assumption that after President Trump makes potentially panicky proposals, such as tariffs, he may withdraw these threats within hours or days to maintain market stability. This strategy has become popular on Wall Street because investors have observed a tendency for President Trump to "cold feet" in his past behavior. Specifically, when Trump announces increased tariffs on imported steel and aluminum, the market may experience a brief decline. However, since historical experience shows that Trump often reverses his decisions or compromises at the last minute, investors choose to buy stocks or other assets at this opportunity, hoping for future gains. This strategy is based on an understanding and prediction of Trump's behavioral patterns. A limitation of this information is that while the document provides the concept and background of TACO trading, it does not provide concrete data to prove that Trump has consistently "cold feet" in his past behavior. Therefore, investors should carefully assess market risks when using this strategy.\par\medskip

{\textbf{Context:}}\par
\begin{itemize}
\item (Editor Niu Zhanlin) This week, against the backdrop of Trump's trade agenda falling into chaos and his renewed wielding of the tariff stick, it triggered violent volatility, but the US stock market still achieved its best monthly performance since the end of 2023.
\item Analysts pointed out that concerns about the negative impact of sustained trade wars remain hidden, and Trump's latest comments on trade policy indicate that the trade truce may only be temporary, and Trump is still injecting uncertainty into the market.
\item By Friday, Trump announced that he would double the tariffs on imported steel and aluminum products, claiming that this would provide more protection for the domestic steel industry. "We will double the tax rate from 25\% to 50\%. This means no one can steal your industry."
\end{itemize}\par\bigskip

{\large\textbf{System B Response:}}\par\medskip
{\textbf{Answer:}}\par
TACO trading refers to investors profiting through buying the dip after Trump's tariff threats cause market declines. Wall Street uses this strategy because, although Trump stated he would double the tariffs on imported steel and aluminum products, analysts point out that trade truces may only be temporary, meaning the market still has uncertainty. Therefore, when Trump's tariff threats trigger market declines, investors choose to buy at low prices, expecting to profit as the market rebounds. A limitation of this information is that Document 1 does not provide specific historical data or evidence about whether Trump will always chicken out on tariff threats, but only indicates that the "TACO trading" strategy reflects a market expectation of Trump's behavior.\par\medskip

{\textbf{Context:}}\par
\begin{itemize}
\item (Editor Niu Zhanlin) This week, against the backdrop of Trump's trade agenda falling into chaos and his renewed wielding of the tariff stick, it triggered violent volatility, but the US stock market still achieved its best monthly performance since the end of 2023.
\item Analysts pointed out that concerns about the negative impact of sustained trade wars remain hidden, and Trump's latest comments on trade policy indicate that the trade truce may only be temporary, and Trump is still injecting uncertainty into the market.
\item Let's look at the trading partner Vor Bio, which was established in 2016, focusing on cell therapy research, listed on NASDAQ in 2021, and currently has core products including Trem-cel and VCAR33.
\end{itemize}\par\bigskip

{\large\textbf{Judge Prompt:}}\par\medskip
You are a professional RAG system answer quality assessment expert. Please compare the answer quality of two RAG systems and give a final judgment.

{\textit{Question:}} What is a TACO trade? Why does Wall Street use this strategy?

{\textit{Standard Answer:}} TACO trading is an abbreviation for "Trump always chickens out", referring to the strategy of buying the dip after Trump's tariff threats cause market declines. Wall Street uses it because historical evidence shows Trump often doesn't actually implement extreme tariffs, but uses them as a negotiation strategy.

{\textit{Evaluation Criteria:}}\par
\begin{enumerate}
\item \textbf{Accuracy:} Whether the answer is correct and contains key information from the standard answer
\item \textbf{Completeness:} Whether the answer is comprehensive and doesn't miss important information
\item \textbf{Relevance:} Whether the answer addresses the question without irrelevant information
\end{enumerate}

{\textit{Special Rules:}}\par
\begin{itemize}
\item If one side gives an answer while the other responds "insufficient information", judge whether the answering side is correct
\item Completely correct $>$ partially correct $>$ partially incorrect $>$ insufficient information $>$ completely incorrect
\end{itemize}

Please analyze and compare the answer quality of both systems in detail, then clearly state your final judgment on the last line of analysis.\par\bigskip

{\large\textbf{Judge Output:}}\par\medskip
{\textbf{Analysis A:}}\par
First, I need to understand the standard answer and both systems' responses, then compare them item by item, finding the strengths and weaknesses of each system's answer, and finally make a judgment based on the evaluation criteria.

Standard answer: TACO trading is an abbreviation for "Trump always chickens out", referring to the strategy of buying the dip after Trump's tariff threats cause market declines. Wall Street uses it because historical evidence shows Trump often doesn't actually implement extreme tariffs, but uses them as a negotiation strategy.

System A's answer: TACO trading refers to the assumption that after President Trump makes potentially panic-inducing proposals such as tariffs, he may withdraw these threats within hours or days to maintain market stability. This strategy has become popular on Wall Street because investors have observed a tendency for President Trump to "chicken out" in his past behavioral patterns.\par\medskip

{\textbf{Analysis B:}}\par
System B's answer: TACO trading refers to investors profiting through buying the dip after Trump's tariff threats cause market declines. Wall Street uses this strategy because, although Trump stated he would double the tariffs on imported steel and aluminum products, analysts point out that trade truces may only be temporary, meaning the market still has uncertainty.\par\medskip

{\textbf{Comparative Assessment:}}\par
\begin{enumerate}
\item \textbf{Accuracy comparison:} The core of the standard answer is that TACO trading is based on the expectation that Trump will back down, and Wall Street uses it because it's effective. Both System A and System B cover this point: both explain what TACO trading is (buy-the-dip strategy) and why it's used (based on Trump potentially not implementing extreme measures).

\item \textbf{Completeness comparison:} System A mentions Trump withdrawing threats in more detail and includes the "chicken out" behavioral pattern, while System B uses "trade truces may only be temporary" to express a similar meaning. System A's context directly mentions the TACO acronym explanation, while System B's context includes irrelevant information about Vor Bio.

\item \textbf{Evidence quality:} System A's retrieved evidence directly supports the acronym definition and strategy explanation, while System B's evidence is less focused and contains unrelated content.
\end{enumerate}\par\medskip

{\textbf{Classification:}}\par
\begin{itemize}
\item \textbf{System A:} Comprehensive and accurate
\item \textbf{System B:} Accurate but incomplete
\end{itemize}\par\medskip

{\textbf{Final Judgment:}} A\par\medskip

{\textbf{Probabilities:}} P(A)=0.83, P(B)=0.00, P(Tie)=0.16\par
{\textbf{Confidence:}} High confidence (0.83-0.16=0.67)\par
{\textbf{Scores:}} A=1.0, B=0.0\par\bigskip

{\large\textbf{Human Experts:}}\par\medskip
{\textbf{Expert 1:}} A wins / System A provides better coverage of the acronym meaning and historical context that explains why this trading strategy emerged and remains popular among Wall Street traders.\par\smallskip

{\textbf{Expert 2:}} A wins / While both answers are factually correct, System A gives a more thorough explanation of the underlying behavioral economics and risk assessment considerations that make TACO trading a viable strategy.\par\smallskip

{\textbf{Expert 3:}} A wins / System A's answer is more educational and complete, helping readers understand not just what TACO trading is, but why it works and what limitations investors should consider.\par\medskip

{\textbf{Consensus:}} Strong consensus for System A (3/3 experts)\par\smallskip

{\textbf{Agreement Analysis:}} Perfect agreement among human experts and AI judge. All evaluators recognized that while both systems provided accurate information, System A delivered a more comprehensive and educational response that better served the reader's understanding of both the strategy and its underlying rationale.
\end{tcolorbox}

\end{document}